\documentclass[letterpaper]{article}
\usepackage{aaai}
\usepackage{times}
\usepackage{helvet}
\usepackage{courier}
\usepackage{latexsym}
\usepackage{url}
\usepackage{graphicx}
\usepackage{dirtytalk}
\usepackage{booktabs} 
\usepackage{caption} 
\usepackage{amsmath}
\usepackage{amsfonts}
\usepackage{algorithm}
\usepackage{algpseudocode}
\usepackage{array,multirow}
\usepackage{xcolor}
\usepackage{hyperref}

\begin{document}
%
\title{Transfer Reward Learning for Policy Gradient-Based Text Generation}

\author{James O' Neill \and Danushka Bollegala \\
Department of Computer Science, University of Liverpool, UK \\
$\mathtt{\{james.o-neill, danushka.bollegala\}@liverpool.ac.uk}$
}

\maketitle
\begin{abstract}
\begin{quote}
Task-specific scores are often used to optimize for and evaluate the performance of conditional text generation systems. However, such scores are non-differentiable and cannot be used in the standard supervised learning paradigm. Hence, policy gradient methods are used since the gradient can be computed without requiring a differentiable objective.
However, we argue that current n-gram overlap based measures that are used as rewards can be improved by using model-based rewards transferred from tasks that directly compare the similarity of sentence pairs. 
These reward models either output a score of sentence-level syntactic and semantic similarity between entire predicted and target sentences as the expected return, or for intermediate phrases as segmented accumulative rewards.
\newline
We demonstrate that using a \textit{Transferable Reward Learner} leads to improved results on semantical evaluation measures in policy-gradient models for image captioning tasks. Our InferSent actor-critic model improves over a BLEU trained actor-critic model on MSCOCO when evaluated on a Word Mover's Distance similarity measure by 6.97 points, also improving on a Sliding Window Cosine Similarity measure by 10.48 points. Similar performance improvements are also obtained on the smaller Flickr-30k dataset, demonstrating the general applicability of the proposed transfer learning method.
\end{quote}
\end{abstract}

\section{Introduction}
Neural network based encoder-decoder architectures are increasingly being used for conditional text generation given the recent advances in Convolutional Neural Networks(CNNs)~\cite{szegedy2017inception} and Recurrent Neural Networks (RNN)~\cite{hochreiter1997long} that use internal gating mechanisms to preserve long-term dependencies~\cite{hochreiter1997long}, showing impressive results for density estimation (i.e language modelling) and text generation.
These encoder-decoder networks are usually trained end-to-end using Maximum Likelihood (ML) training with full supervision (i.e the model learns explicitly from expert demonstrations). This is also referred to as \textit{teacher forcing}~\cite{sutton2000policy}. 

Typically for text generation tasks, such as image captioning, CNNs~\cite{lecun1995convolutional} encode an image, which is passed to the beginning of the decoder RNN language model via a linear map from the image encoding. The decoder policy $\hat{\pi}$ then performs a set of actions given an \textit{expert} policy $\pi^{*}$ for $T$ time steps (e.g human captions). 
\newline
However, ML can act as a poor surrogate loss for a task-specific score we are interested in and evaluate on (e.g BLEU). Moreover, these scores are non-differentiable and hence cannot be used in the standard supervised learning paradigm.
\newline
Deep Reinforcement Learning (DRL) can be used to optimize for task scores as rewards~\cite{bahdanau2016actor,zhang2017actor}, for the purposes of higher quality text generation e.g n-gram overlap based metrics such as ROUGE-L and BLEU.
\newline
Actor-critic networks in particular have shown State of The Art (SoTA) results for image captioning~\cite{bahdanau2016actor,zhang2017actor}. These models use a policy network that produces actions which are evaluated by a value network that output scores given both actions and targets as input for each state. The critic value prediction is then used to train the actor network, assuming the critic values are exact (pre-training the critic is often necessary).
However, parameter-free measures such as BLEU and ROUGE-L do not correlate well with human judgements when compared to using word and sentence-based embedding semantic similarity evaluation measures between predicted and target sequences, moreover the frequency distribution of generated captions is significantly different to human captions~\cite{cui2018learning}. 
\newline
We argue that n-gram overlap based measures in DRL models can be improved using model-based reward estimators that are transferred from sentence similarity models that directly learn from the manually annotated similarity of paired sentences.
\newline
In the context of DRL, we view this type of transfer learning as generating the environment of a target task given a source model that implicitly learns relationships on a pairwise source task (i.e sentence similarity). Additionally, the reward is continuous everywhere and out-of-vocabulary terms do not hinder the TRL model's ability to estimate return since sentence similarity can still be inferred even when $\langle$unk$\rangle$ tags are used at test time to replace tokens we have not seen at training time. We are further motivated by the fact that transfer learning for text generation has already been successfully demonstrated using pretrained ImageNet CNN encoders~\cite{krizhevsky2012imagenet}. However, transfer learning with respect to the decoder of DRL-based encoder-decoder models has been unexplored until this point.  
\newline
We also note that for the common use of DRL in games and robotics, transfer learning is often made difficult since the environments and dynamics are often distinctly different from one another (e.g games usually do not have the same states, actions, transition probabilities and rewards i.e Markov Decision Process (MDP)). In contrast, the MDP for natural language is defined by the vocabulary used for a given corpus. Thus, given a sufficient amount of text the MDP for all corpora converge. Hence, transfer learning becomes easier which is not typical for robotics and games.



\paragraph{Contributions}
This paper proposes to transfer pairwise models that have been trained to learn a similarity score between various universal phrase and sentence representations. These models are trained on a set of sentence-pairwise learning problems such as sentence similarity and natural language inference (NLI).
\newline
Herein, we refer to this as \textit{Transferable Reward Learning} (TRL), a method that incorporates model-based reward shaping to improve task-specific scores in relation to semantic similarity as a measure of language generation quality. We baseline both unsupervised and supervised TRL models against ML training and previous actor-critic models with model-free rewards such as BLEU and ROUGE-L. To our knowledge, this is the first work that focuses on learning to transfer model-based reward in sequence prediction.

\section{Related Work}\label{sec:rw}

\subsection{DRL-based Conditional Text Generation}
~\citeauthor{zhang2017actor} have previously used actor-critic sequence training for image captioning using a token-level advantage and value function, achieving SoTA on MSCOCO at that time. In contrast, TRL can evaluate return both on token and sentence level learned from human judgment similarities between sentences. 

~\citeauthor{ranzato2015sequence} proposed Mixed Incremental Cross-Entropy Reinforce (MIXER) which uses REINFORCE~\cite{williams1992simple} for text generation with a mixed expert policy (originally inspired by DAgger~\cite{ross2014reinforcement}), whereby incremental learning is used with REINFORCE and cross-entropy (CE). During training, the policy gradually deviates from $\pi^{*}$ provided using CE, to using its own past predictions. 

~\citeauthor{rennie2017self} propose Self-Critical Sequence Training (SCTS) which extends REINFORCE by using the test-time model predictions to normalize the reward. This avoids the use a baseline to normalize the rewards and reduce variance, while mitigating exposure bias. 

Reward Augmented ML (RAML)~\cite{norouzi2016reward} combine conditional log-likelihood and reward objectives while showing that highest reward is found when the conditional is proportional to the normalized exponentiated rewards, referred to as the payoff distribution.

~\citeauthor{ren2017deep} have defined the reward as the embedding similarity between both images and sentences that are projected into the same embeddding space, instead of similarity of embeddings corresponding to predicted and target tokens alone. To the best of our knowledge, it is the only other method that uses a continuous reward signal from an embedded space. We can consider this to be an embedding measure that is model-free. In the context of this work, we also consider a model-free sentence embedding similarity measure (see \autoref{sec:trl_method}) in our AC model.

\subsection{Sentence Representations}
Given that our main contribution is the adaptation of TRL pairwise models to model rewards (i.e sentence-level similarity between predicted and target caption), we briefly introduce the relevant SoTA that we consider in our experiments.
~\citeauthor{kiros2015skip} presented Skipthought vectors which are formed using either a bidirectional or unidirectional Long Short Term Memory (LSTM) encoder-decoder that learns to predict adjacent sentences from encodings of the current sentence, which is a fully unsupervised approach. 
\newline
In contrast to Skipthought, Conneau et al. (2017) propose InferSent which is a supervised method to learn sentence representations from natural language inference data~\cite{bowman2015large}. InfeSent outperforms unsupervised sentence representations such as Skipthought on various sentence-pair tasks. We too include this approach in our experiments as the second TRL model to estimate accumulative rewards. 
O' Neill and Bollegala (2018) proposed contextualized embeddings by learning to reconstruct a weighted combination of multiple pretrained word-embeddings as an auxiliary task, acting as a regularizer for the main task.


\subsection{Learning Reward Functions}
Apprenticeship learning~\cite{abbeel2004apprenticeship} has also focused on learning the reward function from expert demonstrations, albeit in the context of robotics. ~\citeauthor{christiano2017deep} proposed to learn to play Atari using deep RL from human preferences. This is analgous to learning from similarity scores that are used as labels for pairwise learning between sentence representations. We share a similar motivation in that learning from demonstrations (i.e reference captions) can be difficult, particularly when there is many permutable demonstrations that lead to a similar goal (i.e many semantically and grammatically correct ways to describe what is in an image). By learning a pairwise-model learned from human preference scores (e.g sentence similarity scores) between trajectories, we can model the reward.

\section{Methodology}

\subsection{Image Caption Setup}
For an image $I$ there is a corresponding caption sequence $Y$ that contains tokens $Y = (y_1, .., y_T)$ and $y \in \mathcal{V}$ where $\mathcal{V}$ is the vocabulary. $f_{\omega}$ encodes an image $I$ into a hidden state $z$ that is then passed to a Recurrent Neural Network (RNN) decoder $f_{\psi}$ that generates a predicted sequence $\hat{Y}$ which is then evaluated with a task-specific score $R(Y, \hat{Y})$. In the DRL setting, we can consider the problem as a finite Markov Decision Process (MDP) where each word $w \in \mathcal{V}$ is considered a state $s \in S$ and a prediction $\hat{y}_t$ is considered as an action $\pi_{\theta}(a_t|s_t)$ in action space $a \in \mathcal{A}$ with probability $p_{\pi_{\theta}}(a_{t}|s_{t})$. The environment then issues a discounted return $g = \sum_{t \in T}\gamma_t r_{t}$ where the discount factor $\gamma \in [0, 1]$, after receiving the set of actions and the objective is then to maximize the total expected return $G$.
We use an actor-critic model~\cite{barto1983neuronlike} as the basis of our experiments with a ResNet-152 encoder~\cite{he2016deep} and an LSTM network.

\subsection{Policy Gradient Training}
We define a policy network $\pi_{\theta}$ as an encoder-decoder architecture that encodes an image $I_s \in \mathbb{R}^{m \times n}$ as $h_s \in \mathbb{R}^{n}$ through the Resnet-152~\cite{he2016deep} CNN-based encoder and a linear projection $W_s \in \mathbb{R}^{d\times n}$ shown in \autoref{eq:resnet}. This is then concatenated with the embedding $x_t \in \mathbb{R}^{m}$ corresponding to the input word $w_t \in \mathbb{Z}$, which forms state $s_t = x_t \oplus h_s$ where $\oplus$ denotes concatenation. 
This LSTM decoder takes $(s_t, h_{t-1})$ as input, as shown in \autoref{eq:enc_dec}, omitting $t=0$ where $h_0$ is used instead. Therefore the policy network parameters include the ResNet-152 parameters $\omega$, the linear projection $W_s$, the LSTM parameters $\psi$ and decoder projection layer $W_t$, hence $\theta:= \{\omega, W_s, \psi, W_t\}$.
The predictions for a given sequence length of $T=|Y|$, predictions are defined as $\hat{Y} = \{a_1,..a_t,.. a_T\}$ where the action space $a_t \in \mathcal{A}$ is defined by the vocabulary $w \in \mathcal{V}$ and the targets $Y = \{w_1,.,w_t,... w_T\}$. 

\begin{align}
  h_s = W_s \text{ResNet-152}(I_s) \label{eq:resnet} \\
  h_t = \text{LSTM}(s_t, h_{t-1}) \label{eq:enc_dec} \\
  p_{\pi_{\theta}}(s_t) = \phi(W_t \cdot h_t) \\
  \pi_{\theta}(a_t|s_t) = p_{\pi_{\theta}}(a_t|s_t)
\end{align} 

\paragraph{Value Function Approximation}

For Value Function Approximation (VFA), the gradient of the expected accumulative discounted reward is typically estimated as $\mathbb{E}[\sum_{t=0}^{T}\gamma_{t}r_{t}|a_{t+1},..a_{T}]$, for reward $r_t$ at time $t$ for an $l$-step return.

We then use a critic network to estimate the state-value function that takes the policy $\pi$, actions $a_{t:t+l} \in \hat{Y}$, parameterized rewards $r^{\vartheta}_{t+1:t+l}$ and compute the expected return $V^{\pi}(s_t)$ from state $s_t$ for $l$ steps. This is given as the expectation over the sum of discounted rewards expressed as \autoref{eq:value_est} where rewards $r^{\vartheta}$ are issued by our proposed TRL model-based reward with frozen parameters $\vartheta$ and $\gamma$ not used as discounted rewards are not applicable in the TRL.

\begin{equation}\label{eq:value_est}
 V^\pi(s_t) = \mathbb{E}\big[\sum_{t=0}^{l}r^{\vartheta}_{t+1}|a_{t+1},..a_{l} \big]
\end{equation}

\paragraph{Advantage Function Approximation}
Above, we considered using $V^{\pi}(s_t)$ to estimate $g$. However, training the value network from scratch can result in high-variance in the gradient and therefore poor convergence. Similarly to~\citeauthor{zhang2017actor}, we use the Advantage Function Approximator (APA) $A^{\pi}$ to reduce the variance in gradient updates.
This is achieved using temporal-difference $\lambda$ (TD-$\lambda$) learning as shown in \autoref{eq:afa} where the Q-function is $Q^{\pi}(s_t, a_t) = \mathbb{E}_{s_{t+1}:T,\\a_{t+1}:T} \Big[\sum_{i=0}^{T}r_{t+l}\Big]$, for N step expected return $G^{i}_{t}$ and $\lambda$ is the trace decay weight $0 \le \lambda \le 1$ (larger $\lambda$ assigns more credit assigned to distant rewards).

\begin{equation}\label{eq:afa}
\begin{gathered}
A^{\pi}(s_t, a_{t+1}) = Q^{\pi}(s_t, a_{t+1}) - V^{\pi}(s_t) \\ = (1-\lambda)\sum_{i=1}^{N}G^{i}_{t} - V^{\pi}(s_t)  
\end{gathered}
\end{equation}
 
The gradient of the policy network can then be rewritten as \autoref{eq:policy_acq_grad}. Here, the trace decay parameter $0 < \lambda < 1$, in our experiments $\lambda=1$ which corresponds to Monte-Carlo and means that large traces are also assigned to distant states and actions. In the context of image captioning, it is typical that the episodes are short ($T < 30$) and hence it is feasible.

\begin{equation}\label{eq:policy_acq_grad}
\begin{gathered}
 G = \mathbb{E}\Big[\sum_{t=0}^{T-1}\big((1-\lambda)\sum_{i=0}^{N}G^{m}_{t} \\
  -V^{\pi}(s_t)\nabla_{\theta}\log \pi_{\theta}(a_{t+1}|s_{t})\Big]
\end{gathered}
\end{equation}

This is achieved by computing the gradient of the log likelihood multiplied by the advantage function $A^{\pi}(s_t, a_{t+1})$ shown in \autoref{eq:grad_pg}. Here, $A^{\pi}(s, a) = Q^{\pi}(s, a) - V^{\pi}(s)$ reduces the variance of the of the gradient by increasing the probability of actions when $A^{\pi}(s_{t}, a_{t+1}) > 0$ and decrease otherwise. 

\begin{equation}\label{eq:grad_pg}
G := \mathbb{E} [\sum^{T-1}_{t=0}A^{\pi}(s_{t}, a_{t+1}) \nabla_{\theta} \log \pi_{\theta} (a_{t+1}|s_{t})]  
\end{equation}

\begin{figure*}
\begin{center}
 \includegraphics[scale=0.5]{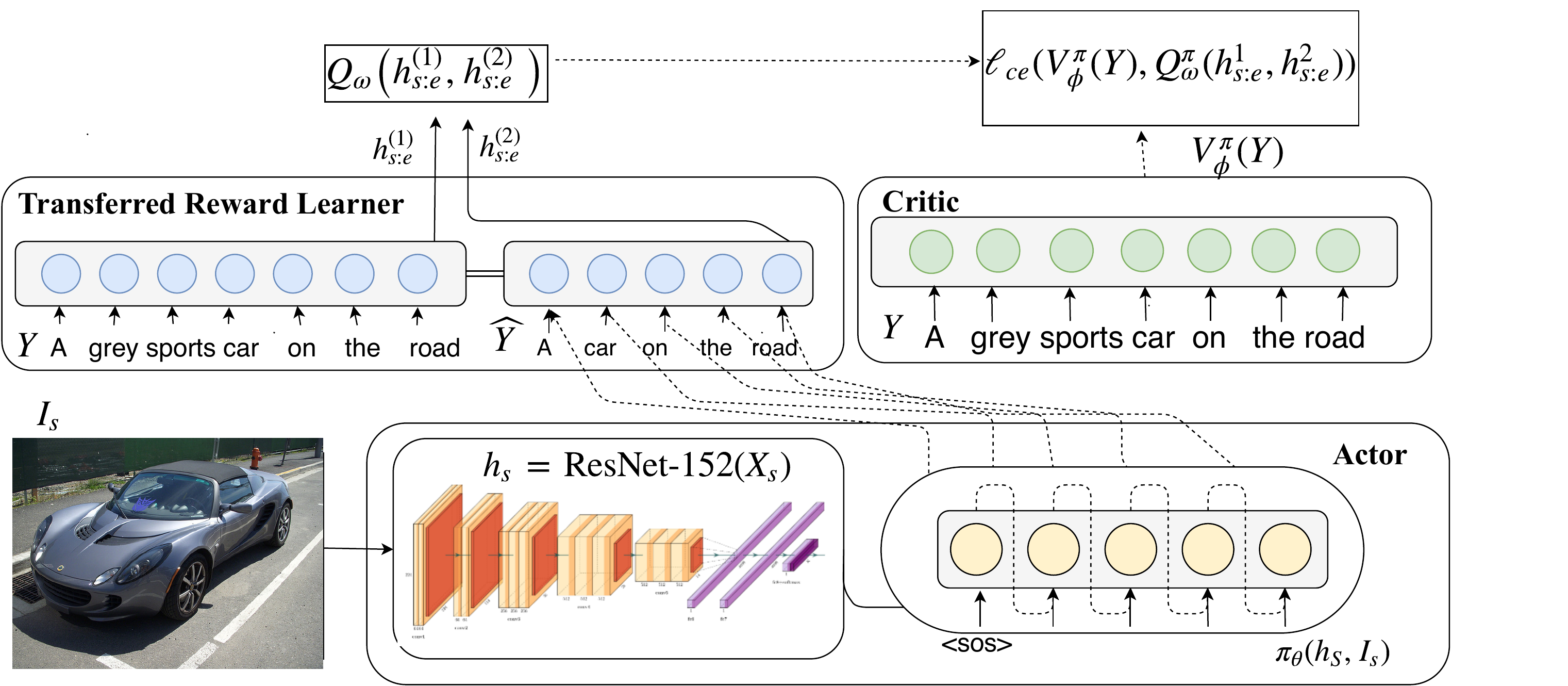}
 \caption{Actor $\pi_{\theta}(\cdot)$ produces actions $\hat{Y}$ given an encoded image $h_s$ and caption $X_t$ that are passed to $\text{TRL}(Y, \hat{Y})$ and encoded into $(h_1, h_2)$ respectively and scored with action-value $A^{\pi} = V^{\pi}_{\Theta}(s) - Q^{\pi}(s, a)$}\label{fig:TRL}
\end{center}
\end{figure*}

\subsection{Transfer Reward Learner}\label{sec:trl_method}
We now consider two sentence encoders, as mentioned in \autoref{sec:rw}, as the TRL. We note that, although there has been considerable breakthroughs in recent years for models that could be used for sentence similarity tasks~\cite{devlin2018bert}, these models are too computationally costly to consider for issuing rewards and typically have more parameters than the whole actor-critic network combined. 

Both TRL models that evaluate state-action pairs are denoted as $R_{\vartheta}(s, a)$ where the $\vartheta$ parameters are not-updated as rewards are kept static throughout training. The advantage of this is that we are not restricted to choosing $\lambda=1$, which is used for the sentence-level n-gram overlap measures such as BLEU, ROUGE and CIDEr. The critic can evaluate partially generated sequences and sentence pairs of different length, since they have been trained to learn similarity between sentences of non-equal length. We emphasize at this point that the TRL is not updated for value function estimation in our experiments, this is only carried out for approximating the advantage and value functions. 

\paragraph{InferSent Rewards}
For reward shaping, we use a pretrained sentence similarity model such as InferSent, tuned on SemEval 2017 Semantic Textual Similarity (STS) dataset\footnote{ \url{http://alt.qcri.org/semeval2017/task1/}} consisting of English monolingual pairs that are labelled with a score from 0 (semantic independence) to 5 (semantic equivalence). The scores are scaled from the continuous [0-5] range to [0-1] using a sigmoid $\sigma(\cdot)$ to convert to a probability.

Conneau et al. (2017) use the scoring function in \autoref{eq:infersent_score} between two encoded sentence pairs $(h_1, h_2)$ where $h_1, h_2 \in \mathbb{R}^{d}$, corresponding to two sentences $(\mathcal{S}_1, \mathcal{S}_2)$. We also use this scoring function for the pretrained InferSent model.


\begin{equation}\label{eq:infersent_score}
\phi\big([h_1, h_2, |h_1-h_2|, h_1\cdot h_2]\cdot W + b\big)
\end{equation}

The model is a Bidirectional-GRU (or BiLSTM) with max (or mean) pooling, as in \autoref{eq:infer_mod} where $g$ represents the pooling function and $e_i$ is the embedding corresponding to word $x_i$.

\begin{equation}\label{eq:infer_mod}
h = g_{\text{max-pool}}\big([\overrightarrow{\text{\small{GRU}}}(e_1, ..,e_T), \overleftarrow{\text{\small{GRU}}}(e_1, ..,e_T)]\big)    
\end{equation}

We also use the self-attentive variation in \autoref{eq:self_att_infer_mod}, where the max-pooling operation $g$ is replaced with self-attention that produces a weighted average where $g_{\text{avg}}(\cdot)$ sum the weights to 1 $\forall t \in T$. Hence, attention focuses on the hidden states of important tokens prior to using the scoring function.

\begin{equation}\label{eq:self_att_infer_mod}
    h = \sum_{t=1}^{T}g_{\text{avg}}(\tanh(W h_t + b)) h_t
\end{equation}

\paragraph{Skipthought Rewards}
We also consider the Skipthought vectors as unsupervised sentence representations, which have shown competitive performance on sentence-level pairwise tasks~\cite{kiros2015skip}. This allows us to compare against the supervised InferSent model. Similarly to the InferSent model, we use the same scoring function as in \autoref{eq:infersent_score}.

\paragraph{Critic Loss} 

Prior work in this area used an $\ell_2$ loss between $\pi_{\theta}(a|s)$ policy values and the critic scores $Q^{\pi}(s, a)$~\cite{zhang2017actor}. In our experiments, we found a KL-divergence loss outperformed an $\ell_2$ loss.

We minimize the KL divergence between the $[0, 1]$ normalized $\tilde{Q}^{\pi}(s_t, a_{t+1})$ and $\tilde{V}^{\pi}_{\Theta}(s_t) $ which corresponds to minimizing the cross-entropy loss when ignoring $\mathcal{H}\big(\tilde{Q}^{\pi}(s_t, a_{t+1})\big)$ that does not rely on $\Theta$. The logit penalizes values that are much higher than the baseline more than the $\ell_2$ loss but show a larger gap between small value improvements over the baseline. 

\begin{gather}\label{eq:kl_loss}  
  \ell_{ce} = \mathcal{D}_{\text{KL}}\Big(\tilde{Q}^{\pi}(s_t, a_{t+1})|| \tilde{V}^{\pi}_{\Theta}(s_t) \Big) \nonumber \\ + \mathcal{H}\big(\tilde{Q}^{\pi}(s_t, a_{t+1})\big)
\end{gather}

This has the effect of stabilizing the critic network over consecutive iterations, as the critics gradient updates are not as large, stabilizing the training of the critic and subsequently ensuring the difference in the policy network $\pi_{\theta}(a_t|s_t)$ is less drastic between iterations.

\section{Experimental Setup}

\subsection{Dataset Details}
We use the Microsoft \textit{Common Objects in Context} (MSCOCO) 2014 image captioning dataset proposed by~\citeauthor{lin2014microsoft}, which is a de-facto benchmark
for image captioning. The dataset includes 164,062 images (82,783 training images, 40,504 validation images, and 40,775 test images) with 5 manually labelled captions per image of 80 object categories and 91 \textit{stuff} categories. Each image is paired with as least five manually annotated captions.
\newline
We also use the smaller Flickr-30k~\cite{young2014image} dataset, which contains 30k images with 150k corresponding captions, which also includes a constructed denotation graph that can be used to define denotational similarity, giving more generic descriptions through lexical and syntactic operations.

\subsection{Training Details}
As mentioned before, we use the ResNet-152~\cite{he2016deep} classifier trained on ImageNet as our encoder. The reported experimental results are that of a 2-hidden layer LSTM decoder~\cite{hochreiter1997long} network, with embedding input size and hidden layer size of $|e|=|h|=512$. For both MSCOCO and Flickr30k a mini-batch size of $|x_{sub}| = 80$ and use adaptive momentum ($\mathtt{adam}$)~\cite{kingma2014adam} for stochastic gradient descent (SGD) optimization for the LSTM decoder while, as mentioned, the image encoder is not updated for our experiments. 

Training both actor and critic networks from scratch is difficult, or more generally for related policy gradient algorithms since it is often the case the reward signal leads to high variance in the gradient updates, particularly in the early phase of training where the parameters $\theta$ and $\phi$ are initialized randomly.
\newline
Therefore, in all our experiments we pre-train the actor and critic networks (similarly to ~\citeauthor{ren2017deep}) for 5 and 7 epochs respectively using by minimizing the cross entropy loss $-\sum_{t=1}^{T}\log p_{\theta}(a_t|s_t)$ for both the actor network and the critic network (as mentioned we also consider CSP loss). After the actor is pre-trained, the critic network is passed sampled actions from the fixed pre-trained actor and updated accordingly. After this initial phase, we then begin training both actor and critic together.

\begin{table}
\centering
\captionsetup{justification=centering, margin=0cm}
\resizebox{1.\linewidth}{!}{%
\begin{tabular}{l|cccc|c}
\toprule[2.pt]

\textbf{Methods} & \multicolumn{1}{c}{B1} & \multicolumn{1}{c}{B2} & \multicolumn{1}{c}{B3} & \multicolumn{1}{c}{B4} & \multicolumn{1}{|c}{PPL} \\
\midrule
\textbf{NeuralTalk}~\citeauthor{karpathy2015deep} & 0.57 & 0.37 & 0.24 & 0.16 & - \\
\textbf{Mind’s Eye}~\citeauthor{chen2015mind} & - & - & - & 0.13 & 19.10 \\
\textbf{NIC}~\citeauthor{vinyals2015show} & 0.66 & - & - & - & - \\
\textbf{LRCN}~\citeauthor{donahue2015long} & 0.59 & 0.39 & 0.25 & 0.17 & - \\
\textbf{m-Rnn-AlexNet}~\citeauthor{mao2014deep} & 0.54 & 0.36 & 0.23 & 0.15 & 35.11 \\
\textbf{m-Rnn-VggNet}~\citeauthor{mao2014deep} & 0.60 & 0.41 & 0.28 & 0.19 & 20.72 \\
\textbf{Hard-Attention} ~\citeauthor{xu2015show} & 0.67 & 0.44 & 0.30 & 0.20 & - \\
\midrule
~\citeauthor{liu2017attention} & \\
\textbf{Implicit-Attention} & - & - & 0.29 & 0.19 & - \\
\textbf{Explicit-Attention} & - & - & 0.29 & 0.19 & - \\
\textbf{Strong Sup} & - & - & 30.2 21.0 & 0.19 & - \\
\midrule
~\citeauthor{wu2018image} & \\
\textbf{Att-GT+LSTMz} & 0.78 & 0.57 & 0.42 & 0.30 & 14.88 \\
\textbf{Att-SVM+LSTM} & 0.68 & 0.49 & 0.33 & 0.23 & 16.01 \\
\textbf{Att-GlobalCNN+LSTM} & 0.70 & 0.50 & 0.35 & 0.27 & 16.00 \\
\textbf{Att-RegionCNN+LSTM} & 0.73 & 0.55 & 0.40 & 0.28 & 15.96 \\

\bottomrule[2pt]
\end{tabular}%
}
 \caption{SoTA Methods for Flickr30k}
 \label{tab:flickr_sota}
\end{table}

\begin{table*}[ht]
\centering
\captionsetup{justification=centering, margin=0cm}
\resizebox{1.\linewidth}{!}{%
\begin{tabular}{cc|cc|cc|cc|cc|cc|cc}
\toprule[2.pt]

& \textbf{Flickr-30k} & \multicolumn{2}{c|}{ROUGE-L} & \multicolumn{2}{c|}{BLEU2} & \multicolumn{2}{c|}{BLEU3} & \multicolumn{2}{c|}{BLEU4} & \multicolumn{2}{c|}{WMD} & \multicolumn{2}{c}{COS} \\

\midrule

& & Val. & Test & Val. & Test & Val. & Test & Val. & Test & Val. & Test & Val. & Test \\

\midrule
\midrule

\parbox[t]{2mm}{\multirow{3}{*}{\rotatebox[origin=c]{90}{\textbf{ \small{Baseline}}}}}


& \multicolumn{1}{l|}{\textbf{ML}} & 33.09 & 31.46 & 67.52 & 65.20 & 42.12 & 40.68 & 27.81 & 26.87 & 65.35 & 63.62 & 60.14 & 59.22 \\

& \multicolumn{1}{l|}{\textbf{BLEU}} & 35.68 & 32.93 & 70.03 & 70.39 & 48.65 & 48.45 & 30.88 & 30.79 & 73.99 & 72.70 & 66.22 & 66.10 \\

& \multicolumn{1}{l|}{\textbf{ROUGE-L}} & 36.47 & 33.67 & 68.98 & 68.55 & 47.55 & 47.14 & 31.68 & 31.56 & 71.94 & 70.08 & 65.55 & 62.42\\


\midrule 

\parbox[t]{2mm}{\multirow{3}{*}{\rotatebox[origin=c]{90}{\textbf{ \small{Our}}}}}

& \multicolumn{1}{l|}{\textbf{WMD}} & 31.61 & 30.76 & 70.06 & 68.05 & 46.24 & 44.23 & 29.78 & 28.94 & 76.59 & 73.40 & 69.83 & 68.73 \\

& \multicolumn{1}{l|}{\textbf{InferSent}} & 30.08 & 29.29 & 69.48 & 68.76 & 44.76 & 43.96 & 28.79 & 28.70 & \textbf{\emph{78.28}} & \textbf{\emph{77.01}} & \textbf{\emph{71.23}} & \textbf{\emph{69.16}} \\

& \multicolumn{1}{l|}{\textbf{Skipthought}} & 26.22 & 25.36 & 70.01 & 67.43 & 43.41 & 42.20 & 29.23 & 27.81 & \textbf{\emph{76.50}} & \textbf{\emph{75.18}} & \textbf{\emph{72.63}} & \textbf{\emph{68.60}} \\

\bottomrule[2pt]
\end{tabular}%
}
  \caption{Flickr30k Results for ML, Actor-Critic and our proposed TRL AC Models (using a beam width of 5)}
 \label{tab:flickr30k_results}
\end{table*}

\subsection{Embedding Similarity Evaluation}

\paragraph{Word Mover's Distance Sentence Similarity}
We also include WMD~\cite{kusner2015word} for measuring semantic similarity between $\ell_2$ normalized embeddings associated with predicted and target words. Word-level embedding similarities offer a faster alternative to model-based sentence-level evaluation, hence we include it for our experiments.

To align WMD with word overlap metrics, we also include the penalization terms such as the brevity penalties used in BLEU~\cite{papineni2002bleu}, as shown in \autoref{eq:aug_wmd}. Here, $\gamma$ is the similarity measure, the length ratio $\text{lr} = |Y|/|\hat{Y}|$~\cite{shi2016neural} and the brevity penalty $\text{bp}= \min \big(\exp(1-1/lr), 1\big)$ which penalizes shorter length generated sentences.

\begin{gather}\label{eq:aug_wmd}
  s = \sigma \Bigg(\text{BP} \cdot \gamma_{\text{wmd}}\big(E_{\hat{Y}}, E_Y\big)\Bigg)
\end{gather}


\paragraph{Sliding Kernel Cosine Similarity} 

We also considered decayed $k$-pairwise cosine similarity where $k$ is a sliding window span that compares embeddings corresponding to n-gram groupings with a decay factor $\gamma \in [0, 1]$ that depends on the distance such that $\gamma_{(i, j)} = d(Y_i, Y_j)/k \; \forall i, j \in T$. In the below case we use the kernel shown in \autoref{eq:kernel} where $i$ is the index corresponding to $y \in Y$ and j for $\hat{Y}$
respectively.

\begin{equation}\label{eq:kernel}
   \gamma = \exp(-||i - j||)
\end{equation}

This allows for any mis-alignments between sentences, as some may be shorter than others. There are $T/k$ window spans, therefore we multiply the $k/T$ by the brevity penalty.

\begin{gather}\label{eq:window_cos_similarity}
  s_{kcos} = \sigma\Big(\frac{k}{T}\text{BP}\sum_{i=1}^{T}\sum_{j=i-k}^{t+k} \gamma_{(i, j)}cos\big(E_{Y_{i}}, E_{\hat{Y}_{j}}\big)\Big) \nonumber\\
   s.t, \quad t \leq i \leq T-k 
\end{gather}

\section{Results}

\begin{table*}[ht]
\centering
\captionsetup{justification=centering, margin=0cm}
\resizebox{1.\linewidth}{!}{%
\begin{tabular}{lc|cc|cccccccc|cccc|cccc}
\toprule[2.pt]

& \textbf{MSCOCO} & \multicolumn{2}{c|}{ROUGE-L} & \multicolumn{2}{c|}{BLEU1} & \multicolumn{2}{c|}{BLEU2} & \multicolumn{2}{c|}{BLEU3} & \multicolumn{2}{c|}{BLEU4} & \multicolumn{2}{c|}{CIDEr} & \multicolumn{2}{c|}{METEOR} & \multicolumn{2}{c|}{WMD} & \multicolumn{2}{c}{COS} \\


\midrule

& & Val. & Test & Val. & Test & Val. & Test & Val. & Test & Val. & Test & Val. & Test & Val. & Test & Val. & Test & Val. & Test \\

\midrule
\midrule

\parbox[t]{4mm}{\multirow{9}{*}{\rotatebox[origin=c]{90}{\textbf{SoTA}}}}

& \textbf{MIXER}~\cite{ranzato2015sequence}  & - & 53.8 & - & - & - & - & - & - & - & 30.9 & - & 101.9 & - & 25.5 & - & - & - & - \\

& \textbf{MIXER-BCMR}~\cite{ranzato2015sequence}  & - & 53.2 & - &  72.9 & - &  55.9 & - & 41.5 & - & 30.6 & - & 92.4 & - & 24.5 & - & - & - & - \\

& \textbf{PG-BCMR}~\cite{ranzato2015sequence}  & - & 55.0 & - & 75.4 & - & 59.1 & - & 44.5 & - & 33.2 & - & 101.3 & - & 25.7 & - & - & - & - \\

& \textbf{SPIDEr}~\cite{DBLP:journals/corr/LiuZYG016} & - & 54.4 & - & 74.3 & - & 57.9 & - & 43.1 & - & 31.7 & - & 100.0 & - & 25.1 & - & - & - & - \\

& \textbf{RAML @ $\tau=0.9$}~\cite{ma2017softmax}  & & - & - & - & - & - & - & - & - & 27.6 & - & - & - & - & - & - \\

& \textbf{VSE}@$\lambda=0.4$~\cite{ren2017deep} & - & 52.5 & - & 71.3 & - & 53.9 & - & 40.3 & - & 30.4 & - & 93.7 & - & 24.7 & - & - & - & -\\

& \textbf{TD-AC}~\cite{zhang2017actor}  & - & \textbf{\emph{55.4}} & - & \textbf{\emph{77.8}} & - & \textbf{\emph{61.2}} & - & \textbf{\emph{45.9}} & - & \textbf{\emph{33.7}} & - & \textbf{\emph{116.2}} & - & \textbf{\emph{26.7}} & - & - & - & - \\

& \textbf{SCST}~\cite{rennie2017self}  & - & 54.3 & - & 
- & - & - & - & - & - & 31.9 & - & 106.3 & - & 25.5 & - & - & - & -\\

& \textbf{SCST}~\cite{wu2018image}  & - & 54.3 & - & 
- & - & - & - & - & - & 31.9 & - & 106.3 & - & 25.5 & - & - & - & -\\

\midrule
\midrule

\parbox[t]{2mm}{\multirow{3}{*}{\rotatebox[origin=c]{0}{\textbf{Baselines}}}}

& \textbf{ML} & 51.39 & 50.28 & 72.73 & 69.09 & 49.70 & 49.33 & 31.89 & 31.45 & 24.09 & 23.67 & 84.93 & 84.03 & 23.68 & 23.45 & 72.89 & 71.46 & 71.01 & 70.55 \\


& \textbf{BLEU} & 52.75 & 52.01 & 75.91 & 74.17 & 61.34 & 61.72 & 47.91 & 46.58 & 35.09 & 34.57 & 94.46 & 93.41 & 25.55 & 25.27 & 74.38 & 73.09 & 73.39 & 71.86 \\

& \textbf{ROUGE-L} & 56.28 & 55.25 & 72.98 & 72.55 & 51.55 & 50.29 & 37.44 & 35.38 & 32.44 & 31.09 & 95.53 & 95.51 & 25.61 & 25.54 & 73.53 & 72.65 & 73.88 & 72.10 \\


\midrule

\parbox[t]{1mm}{\multirow{3}{*}{\rotatebox[origin=c]{0}{\textbf{Proposed}}}}

& \textbf{WMD} & 51.61 & 52.05 & 73.01 & 72.81 & 52.33 & 52.70 & 39.17 & 38.24 & 32.74 & 30.09 & 99.03 & 98.46 & 27.12 & 27.09 & 79.12 & 78.42 & 80.07 & 78.72 \\

& \textbf{InferSent} & \emph{\textbf{55.46}} & \emph{\textbf{54.25}} & \emph{\textbf{75.58}} & \emph{\textbf{75.02}} & \emph{\textbf{60.40}} & \emph{\textbf{57.16}} & \emph{\textbf{46.24}} & \emph{\textbf{41.68}} & \emph{\textbf{31.93}} & \emph{\textbf{31.24}} &  \emph{\textbf{106.12}} &  \emph{\textbf{105.68}} &  27.31 &  27.18 & \emph{\textbf{82.86}} & \emph{\textbf{80.26}} & \emph{\textbf{83.71}} & \emph{\textbf{82.58}} \\

& \textbf{Skipthought} & 53.02 & 52.71 & 74.49 & 73.61 & 54.54 & 51.08 & 32.28 & 31.02 & 29.78 & 28.59 & 105.56 & 105.08 & \emph{\textbf{27.42}} & \emph{\textbf{27.20}} & 81.95 & 80.21 & 81.25 & 80.63 \\

\bottomrule[2pt]
\end{tabular}%
}
 \caption{MSCOCO Results for ML, Actor-Critic and our proposed TRL AC Models (B=5)}
 \label{tab:coco_results}
\end{table*}

\subsection{Flickr30k Results}

\autoref{tab:flickr_sota} shows the SoTA for image captioning on the Flickr30k dataset, not specific to policy-gradient methods as not all relevant papers include Flickr30k in experiments. Models proposed by \citeauthor{wu2018image} incorporate external knowledge (SPARQL queries over DBpedia knowledge base) for image captioning, hence the increase in BLEU and Perplexity (PPL).

~\autoref{tab:flickr30k_results} compares ML training with previously published actor-critic approaches that use BLEU and ROUGE as the reward signal~\cite{zhang2017actor}. We a beam search of a beam size $B=5$ at test time. The beam search retains $B$ most probable prediction at each timestep and considers the possible next token $w^b_{t+1}$ extensions for a beam $b$ and repeats until timestep $T$, $\forall b \in B$. 

We find that when using only WMD as the reward signal, which is model-free, we find an improvement on semantic similarity measures (i.e WMD and COS). Here, COS refers to the Sliding Kernel Cosine Similarity described in the previous section. Interestingly, we also find WMD improves over ML for word-overlap eventhough the measure does not optimize for a dirac distribution, like ML training. 
\newline
Both TRLs (InferSent and Skipthought) make significant improvements on WMD and COS. Hence, we infer that these TRLs that learn sentence similarity produce semantically similar sentences at the expense of a decrease in word-overlap (expected since the model is not restricted to predicting the exact ground truth tokens). This relaxes the strictness of word-overlap and allows for diversity in the generated captions. Moreover, WMD does not penalize sentence length and thus promotes diversity in caption length. However, as mentioned, we do include brevity penalty in WMD and COS for the purposes of easier comparison to word overlap metrics.

\subsection{MSCOCO Results}
The top of \autoref{tab:coco_results} shows SoTA results for policy-gradient methods based on the best average score on BLEU~\cite{papineni2002bleu}, ROUGE-L~\cite{lin2004rouge}, METEOR~\cite{denkowski2014meteor}, CIDEr~\cite{vedantam2015cider} evaluation metrics. VSE is the aforementioned Visual Semantic Embedding model that uses $\text{TD}(\lambda)$ at $\lambda=0.4$. For SCST, these results are from the test portion of the Karpathy splits~\cite{karpathy2015deep} using CIDEr for optimization. Policy gradient methods have reached near top of the MSCOCO competition leaderboard without using ensemble models.

The lower end shows the results of our proposed models and baselines evaluated on both n-gram overlap based measures and word-level (Cosine) and sentence-level (WMD) embedding based evaluation measures. We find that the largest gap in performance between our proposed TRLs and n-gram overlap metrics (BLEU and ROUGE-L) reward signals are found on the embedding-based evaluation measures.
\newline
For all TRLs (WMD, InferSent, Skipthought) performance consistently improves over ML, BLEU and ROUGE-L when evaluated on WMD and Cosine. This suggests that even though we may not strictly predict the correct word as measured by word-overlap measures, the semantic similarity of sentences is preserved as measured by WMD and Sliding Kernel Cosine Similarity. Furthermore, this results in more diverse text generation as the policy network is not penalized for constructing candidate sentences that do not have high word overlap with the reference captions. 
\newline

TRLs outperform word overlap policy rewards such as BLEU and ROUGE-L on our embedding similarity based metrics. Of the three, we find the InferSent TRL to outperform the other two, with the unsupervised Skipthought TRL being competitive for all metrics. We also see results are competitive to the SoTA. We find similar findings for TRL models evaluated on CIDEr and METEOR.  

\begin{figure}[ht]
\begin{center}
 \includegraphics[scale=0.37]{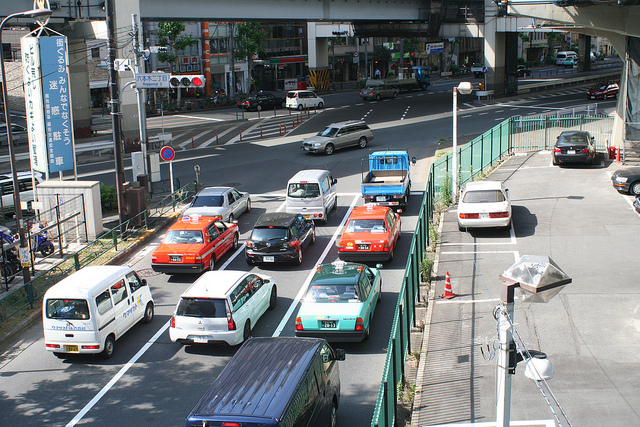}
\vspace{.3cm}

\resizebox{1.\linewidth}{!}{%

\begin{tabular}{|c|c|}
\hline
Human  & \multicolumn{1}{l|}{\textit{An image of \textcolor{orange}{a cars} driving on the highway}} \\
  & \multicolumn{1}{l|}{\textit{A section of traffic coming to a stop at an intersection.}}\\
 & \multicolumn{1}{l|}{\textit{A bunch of cars \textcolor{orange}{sit at the intersection} of a street.}} \\ 
& \multicolumn{1}{l|}{\textit{This is a picture of traffic on a very busy street.}}\\
& \multicolumn{1}{l|}{\textit{A busy intersection filled with cars in asia.}}\\
\midrule
ML & \multicolumn{1}{l|}{\textcolor{blue}{\textit{an image of a sitting car in traffic}}}\\
AC-BLEU &  \multicolumn{1}{l|}{\textcolor{green}{\textit{A group of cars at an intersection.}}}\\
\midrule
AC-WMD & \multicolumn{1}{l|}{\textcolor{red}{\textit{A group of cars at lights near a traffic intersection.}}}\\
AC-Skipthought &  \multicolumn{1}{l|}{\textcolor{red}{\textit{A group of cars near a busy intersection road.}}}\\
AC-InferSent &  \multicolumn{1}{l|}{\textcolor{red}{\textit{A picture of cars stopping near the traffic intersection.}}} \\
\hline
\end{tabular}%
}
 \caption{Qualitative Results on MSCOCO Val. Set}\label{fig:coco_ex}
\end{center}
\end{figure}

\autoref{fig:coco_ex} shows an example of the ground truth captions (Human), ML trained generated caption, a baseline AC trained with BLEU scores and our three proposed alternatives that improve for semantic similarity. We demonstrate the difference between text generated for an image of a traffic jam near an intersection. The example also illustrates that the ground truth itself is imperfect, both syntactically (`..of a cars..') and semantically (`..cars sit at the intersection..'). The TRL will assign lower return in these cases, whereas word-overlap measures do not explicitly penalize how bad the semantic or syntactic differences are between predicted and ground truth sentences.

\section{Conclusion}
We proposed to use pretrained models that are specifically trained on sentence similarity tasks that can be used to issue rewards and to define, optimize and evaluate language quality for neural-based text generation. We find performance on semantic similarity metrics improve over a policy gradient model, namely the actor-critic model, that uses unbiased word overlap metrics as rewards.

The InferSent actor-critic model improves over a BLEU trained actor-critic model on MSCOCO when evaluated on a Word Mover's Distance similarity measure by 6.97 points and 10.48 points on sentence-level cosine embedding metric. Large performance gains are also found for Flickr-30k dataset, demonstrating the general applicability of the proposed transfer learning method. We conclude that model-based task  should be considered for reinforcement learning based approaches to conditional text generation. 

\bibliography{aaai2020}
\bibliographystyle{aaai}

\end{document}